\documentclass[conference, 9pt]{IEEEtran}
\IEEEoverridecommandlockouts
\usepackage{import}
\usepackage{cite}
\usepackage{amsmath,amssymb,amsfonts}
\usepackage{mathtools}
\usepackage{graphicx}
\usepackage{textcomp}
\usepackage{xcolor}
\usepackage{tikz}
\usepackage{multirow}
\usepackage{array}
\usepackage{tabularx}  % To use tabularx
\usepackage{graphicx} 
\usepackage{wasysym}
\usepackage{multirow}
\usepackage{makecell}
\usepackage{graphicx} % For including graphics if needed
\usepackage{array}    % For better table formatting
\usepackage{transparent}
\usepackage[ruled]{algorithm2e}
\def\BibTeX{{\rm B\kern-.05em{\sc i\kern-.025em b}\kern-.08em
    T\kern-.1667em\lower.7ex\hbox{E}\kern-.125emX}}
\newcolumntype{C}[1]{>{\centering\let\newline\\\arraybackslash\hspace{0pt}}m{#1}}

\begin{document}
\newcommand{\blackcircle}[1]{%
  \begin{tikzpicture}[baseline=(char.base)]
    \node[shape=circle, draw, inner sep=0pt, text=white, fill=black] (char) {#1};
  \end{tikzpicture}%
}

\title{GLOVA: Global and Local Variation-Aware Analog Circuit Design with Risk-Sensitive Reinforcement Learning}

% \author{\IEEEauthorblockN{ Dongjun Kim\textsuperscript{*}\IEEEauthorrefmark{*}, Junwoo Park\textsuperscript{1}\IEEEauthorrefmark{*}, Chaehyeon Shin\textsuperscript{2}\IEEEauthorrefmark{*}, Jaeheon Jung\textsuperscript{2}, \\Kyungho Shin\textsuperscript{3}, Seungheon Baek\textsuperscript{3}, Sanghyuk Heo\textsuperscript{3}, Woongrae Kim\textsuperscript{3}, Inchul Jeong\textsuperscript{3}, Joohwan Cho\textsuperscript{3}, and Jongsun Park\textsuperscript{1} }

% % \IEEEauthorblockA{\affiliation{ \textsuperscript{1}Department of Electrical Engineering, Korea University, Seoul, Republic of Korea }\\
% % \affiliation{ \textsuperscript{2}Department of Semiconductor System Engineering, Korea University, Seoul, Republic of Korea }\\
% % \affiliation{\textsuperscript{3}SK hynix, Icheon, Republic of Korea }
% % \thanks{\IEEEauthorrefmark{1}These authors contributed equally to this work.}
% % }}

% \IEEEauthorblockA{{ \textsuperscript{1}Department of Electrical Engineering, Korea University, Seoul, Republic of Korea }\\
% { \textsuperscript{2}Department of Semiconductor System Engineering, Korea University, Seoul, Republic of Korea }\\
% {\textsuperscript{3}SK hynix, Icheon, Republic of Korea }
% \thanks{*These authors contributed equally to this work.}
% }}

\author{
\IEEEauthorblockN{
                    Dongjun Kim\textsuperscript{1}\textsuperscript{*}, 
                    Junwoo Park\textsuperscript{1}\textsuperscript{*}, 
                    Chaehyeon Shin\textsuperscript{2}\textsuperscript{*}, 
                    Jaeheon Jung\textsuperscript{2}, \\
                    Kyungho Shin\textsuperscript{3}, 
                    Seungheon Baek\textsuperscript{3}, 
                    Sanghyuk Heo\textsuperscript{3}, 
                    Woongrae Kim\textsuperscript{3}, 
                    Inchul Jeong\textsuperscript{3}, 
                    Joohwan Cho\textsuperscript{3}, 
                    and Jongsun Park\textsuperscript{1}}

\IEEEauthorblockA{\textsuperscript{1}Department of Electrical Engineering, Korea University, Seoul, Republic of Korea\\
\textsuperscript{2}Department of Semiconductor System Engineering, Korea University, Seoul, Republic of Korea\\
\textsuperscript{3}SK hynix, Icheon, Republic of Korea}

\thanks{*These authors contributed equally to this work.}
}

\maketitle

\begin{abstract}
Analog/mixed-signal circuit design encounters significant challenges due to performance degradation from process, voltage, and temperature (PVT) variations. To achieve commercial-grade reliability, iterative manual design revisions and extensive statistical simulations are required. While several studies have aimed to automate variation-aware analog design to reduce time-to-market, the substantial mismatches in real-world wafers have not been thoroughly addressed. In this paper, we present GLOVA, an analog circuit sizing framework that effectively manages the impact of diverse random mismatches to improve robustness against PVT variations. In the proposed approach, risk-sensitive reinforcement learning is leveraged to account for the reliability bound affected by PVT variations, and ensemble-based critic is introduced to achieve sample-efficient learning. For design verification, we also propose $\mu$-$\sigma$ evaluation and simulation reordering method to reduce simulation costs of identifying failed designs. GLOVA supports verification through industrial-level PVT variation evaluation methods, including corner simulation as well as global and local Monte Carlo (MC) simulations. Compared to previous state-of-the-art variation-aware analog sizing frameworks, GLOVA achieves up to 80.5$\times$ improvement in sample efficiency and 76.0$\times$ reduction in time.
\end{abstract}


\section{Introduction}
Analog/mixed-signal circuit design is an extremely labor-intensive process, relying on human expertise, experience, and intuition. As CMOS technology continues to scale down, it has become increasingly challenging to meet the demand for higher performance and shorter design cycles. Consequently, both industry and academia are driven toward developing advanced design automation tools. Over the years, numerous studies have incorporated machine learning to mitigate reliance on manual design. \cite{genetic1, genetic2} utilize genetic algorithms with deep neural network-based models for optimization. Bayesian optimization (BO) \cite{bo1, bo2, bo3} and reinforcement learning (RL) \cite{gcn_rl, trust_region, robust_analog, pvt_sizing} are also widely used to efficiently explore large design space.

A key challenge in analog circuit design is dealing with performance degradation caused by process, voltage, and temperature (PVT) variations. In particular, process variations during manufacturing introduce substantial silicon mismatches. As shown in Fig. 1, these variations are classified into global mismatches, which occur broadly across the entire wafer, and local mismatches within individual dies \cite{process_variation1, process_variation2}. Voltage variation results from non-idealities of external voltage sources, while temperature variation arises from changes in operating conditions. Since the performance of analog circuits under PVT variations is highly sensitive and unpredictable, addressing these effects is crucial. In industry, where maintaining yield is paramount, this challenge is even more pressing. Therefore, extensive verification across various PVT conditions is conducted to ensure chip reliability, even though it is time-consuming and costly.

Existing design automation tools primarily focus on typical condition and cannot effectively account for diverse variations \cite{genetic1, genetic2, bo1, bo2, bo3, smart_msp}. Although \cite{bo2, smart_msp} introduce PVT corners, they merely test all PVT conditions every iteration, which limits sample efficiency. To tackle this, \cite{robust_analog} employs multi-task RL, treating each PVT corner as a separate task, and improves efficiency by testing only the dominant corners identified through clustering. Nonetheless, using random initial sampling limits both the sample efficiency and success rate of the optimization process. To overcome this, \cite{pvt_sizing} incorporates TuRBO \cite{turbo} for the initial sampling, enhancing optimization efficiency. 

While \cite{bo2, trust_region, robust_analog, pvt_sizing, smart_msp} provide insights into addressing PVT variations, their focus remains primarily on global process corners, leaving the impact of extensive random mismatches both across and within dies underexplored. Although \cite{pvt_sizing} examines transistor-level mismatches, it is limited to a few cases, making it difficult to reflect the comprehensive impact of mismatches. Additionally, the consideration of mismatches is handled separately from the optimization process, requiring additional iterations and reducing overall design efficiency. In terms of verification, ensuring chip reliability necessitates a large number of statistical simulations to address the significant impact of random mismatches in real-world wafers \cite{statistical_simulation}. Nevertheless, prior works have not considered comprehensive verification in the design process. Therefore, it is essential to thoroughly address these factors within the design flow to overcome the challenges in both optimization and verification processes.

\begin{figure}[t]
  \centering
  \includegraphics[width=\linewidth]{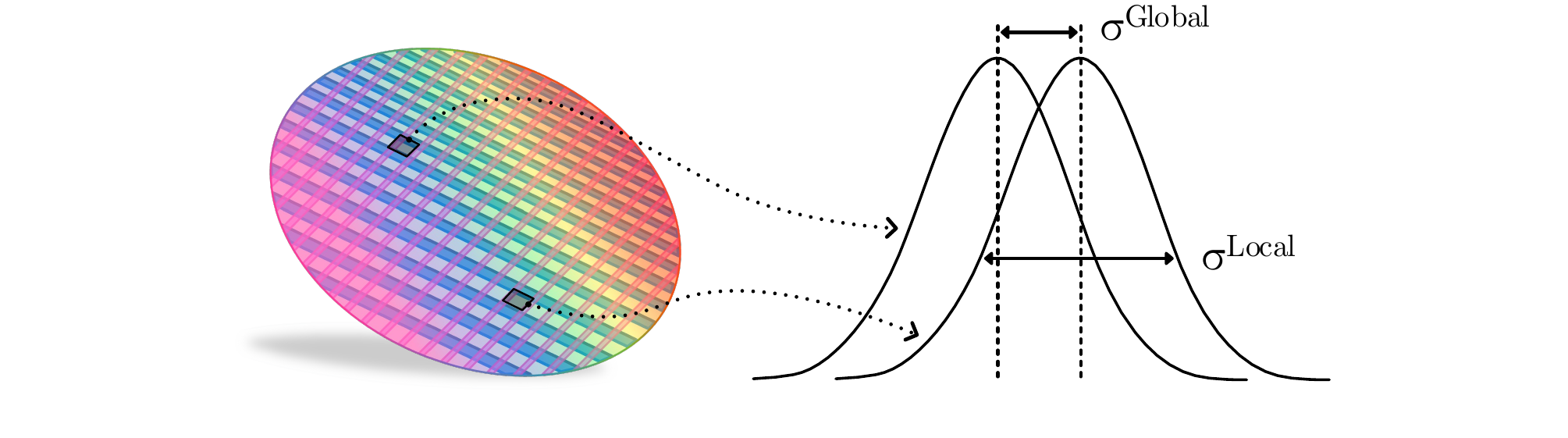}
  \vspace{-0.8cm}
  \caption{Schematic of global (die-to-die) and local (within-die) variations on a wafer \cite{process_variation2}. The median difference between two dies is determined by global variation ($\sigma^{\text{Global}}$), while variations within each die occur around its median due to local variation ($\sigma^{\text{Local}}$).}
  \label{Fig:1}
  \vspace{-0.4cm}
\end{figure}

In this paper, we present GLOVA, an efficient optimization and verification framework for variation-aware analog circuit design automation. Experimental results demonstrate that the proposed GLOVA efficiently and effectively manages diverse PVT conditions and a wide range of mismatch cases compared to previous state-of-the-art approaches.
% \vspace{-\topsep} %이거 없으면 사이에 공백 생김

\begin{figure*}[t]
  \centering
  \includegraphics[width=\textwidth]{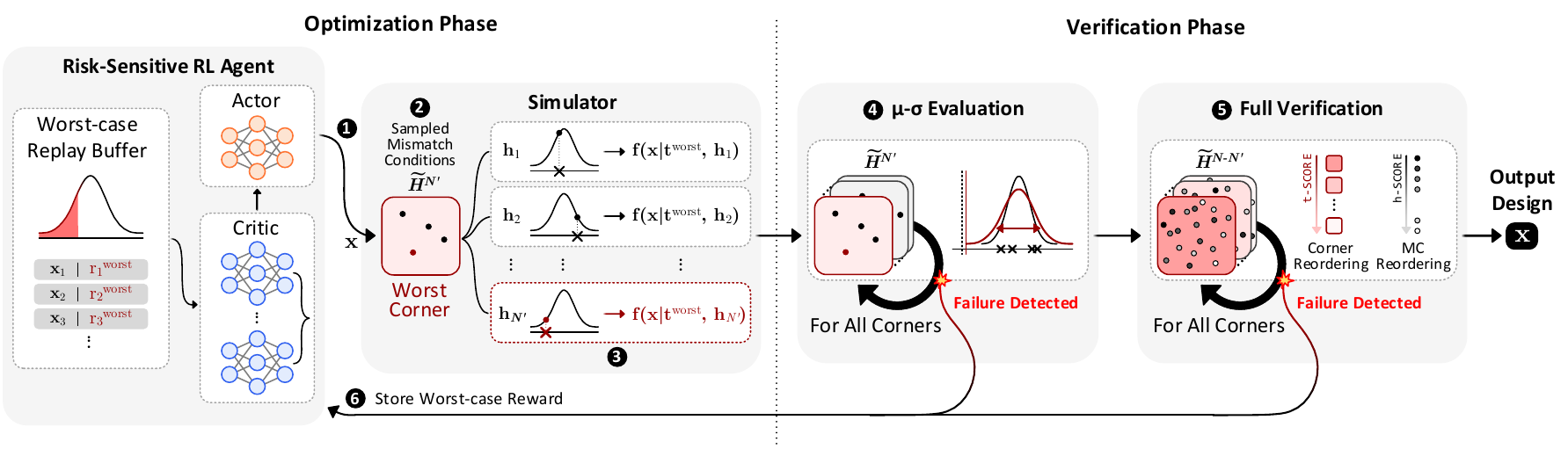}
  % \vspace{-0.8cm}
  \caption{Framework overview of the proposed GLOVA.}
  \label{Fig:2}
  \vspace{-0.3cm}
\end{figure*}

Our contributions are summarized as follows,

\begin{itemize}
    \item We propose an automated optimization and verification framework for PVT-aware analog circuit design, accounting for broad and diverse global and local mismatches via risk-sensitive reinforcement learning.
    \item Ensemble-based critic is proposed to estimate the bound of uncertain worst-case variations, improving sample efficiency and optimization convergence.
    \item $\mu$-$\sigma$ evaluation method is proposed to determine if an optimized design is feasible and whether to further verify it, improving sample efficiency and reducing overall runtime.
    \item Simulation reordering method is proposed, significantly reducing the verification cost of a large number of statistical simulations.
    \item The produced designs are verified through real-world simulation scenarios, including corner simulation, corner with local Monte Carlo (MC) simulation, and corner with global-local MC simulation.
    \item Extensive experimental results demonstrate that GLOVA achieves up to 80.5$\times$ improvement in sample efficiency and 76.0$\times$ runtime reduction compared to state-of-the-art tools.
\end{itemize}

\section{Preliminaries}

\subsection{Simulation Methods for Analyzing PVT Variation Effects}
Corner and Monte Carlo (MC) simulations are commonly used to evaluate analog circuit performance under varying process, voltage, and temperature (PVT) conditions. Corner simulation tests extreme conditions by analyzing predefined PVT corners, such as process corners $\{\mathrm{TT}, \mathrm{SS}, \mathrm{FF}, \mathrm{SF}, \mathrm{FS}\}$, voltage levels $\{0.8\mathrm{V}, 0.9\mathrm{V}\}$, and temperatures $\{-40\thinspace\mathrm{^\circ \/C}, 27\thinspace\mathrm{^\circ \/C}, 80\thinspace\mathrm{^\circ \/C}\}$. Regarding process variation, corner simulation primarily reflects global variation between dies. In contrast, MC simulation statistically assesses process variations by evaluating numerous random parameter combinations. By exploring a large number of combinations, it can probabilistically reveal worst-case scenarios. In practical design, corner simulation and MC simulation can be appropriately combined to experiment with a wide range of mismatch cases under various PVT conditions \cite{hspice_manual}. To ensure robust circuit design, thorough verification using these methods is the fundamental requirement.

\subsection{Risk-Sensitive Reinforcement Learning}
Deep reinforcement learning (RL) is an emerging field used to optimize decision-making in complex environments. In traditional RL, the goal is typically to maximize the expected cumulative reward. In contrast, environments with significant uncertainty introduce unexpected occurrences of negative outcomes that can lead to potential risks. Thus, to focus not only on reward maximization but also on risk management, risk-sensitive RL \cite{risk_original, risk_bnn, risk_survey} aims to balance the expected reward and risk by penalizing deviations from the expected outcomes during simulated rollouts. The objective function $J$ in risk-sensitive RL, referred to as risk-sensitive criterion \cite{risk_survey}, modifies the standard objective to prioritize stability in worst-case scenarios under stochastic uncertainty, and it is expressed as:

\vspace{-0.18cm}
\begin{equation}
\label{Eq:1}
J = \mathbb{E}[R] + \beta \sigma[R]
\end{equation}

\noindent where $\mathbb{E}[R]$ is the expected reward, $\sigma[R]$ is the standard deviation of reward, and $\beta$ is the risk-sensitivity parameter. Adjusting $\beta$ controls the level of risk, with $\beta>0$ for risk-seeking, $\beta<0$ for risk-avoidance, and $\beta=0$ for risk neutrality \cite{risk_bnn}. By modeling and carefully accounting for worst-case scenarios in the optimization process of stochastic and unpredictable systems, this approach helps to minimize failure costs. As a result, it allows desired targets to be reached more efficiently, especially when the cost of failure is high.

\section{Framework Overview of Proposed GLOVA}
\subsection{Problem Formulation}
Our goal is to find a sizing vector ensuring that the circuit’s performance meets the design targets under PVT variations. Therefore, the problem can be formulated as a constraint satisfaction problem:

\vspace{-0.1cm}
\begin{equation}
\begin{aligned}
\text{minimize} &\quad 0 \\
\text{subject to} &\quad \max_{\mathbf{h} \in \widetilde{H}_j^N} \mathcal{F}_i(\mathbf{x} | \mathbf{t}_j, \mathbf{h}) < c_i, \quad i = 1, ..., m, \quad j = 1, ..., k,
\end{aligned}
\end{equation}

\noindent where \( \mathbf{x} \in \mathcal{X}^p \) represents the design solution in a \( p \)-dimensional design space \( \mathcal{X} \), and \( \mathcal{F}_i(\mathbf{x} | \mathbf{t}_j, \mathbf{h}) \) denotes the $i$-th performance metric under the \( j \)-th PVT corner \( \mathbf{t}_j \) and mismatch condition \( \mathbf{h} \). The term $c_i$ represents the $i$-th target constraint. The function \( \mathcal{F}(\mathbf{x} | \mathbf{t}, \mathbf{h}) \) non-linearly maps \( \mathbf{x} \) to performance metric, relying on SPICE simulation to reflect circuit's physical behavior. The PVT corner \( \mathbf{t} \in T \) is an element of the predefined set $T$ of PVT corner conditions. The vector \( \mathbf{h} \in \widetilde{H}^N \) represents a \( r \)-dimensional mismatch condition, and the mismatch condition set \( \widetilde{H}^N \) is obtained by randomly sampling \( N \) times from the distribution over \( \mathbb{R}^r \), as follows:

\vspace{-0.4cm}
\begin{equation}
\begin{aligned}
\mathbf{h}^{(1)} \sim \mathcal{N}(0, \Sigma^{Global}(\mathbf{x})) \\
\widetilde{H}^{N} \coloneqq \{ \mathbf{h}^{(2)}_1, \mathbf{h}^{(2)}_2, \dots, \mathbf{h}^{(2)}_N \mid \mathbf{h}^{(2)} \sim \mathcal{N}(\mathbf{h}^{(1)}, \Sigma^{Local}(\mathbf{x})) \}.
\end{aligned}
\end{equation}

\noindent Eq. (3) outlines a hierarchical process for generating the set $\widetilde{H}^N$. Initially, a global variation sample $\mathbf{h}^{(1)}$ is drawn from a normal distribution $\mathcal{N}\big(0, \Sigma^{\text{Global}}(\mathbf{x})\big)$, where $\Sigma^{\text{Global}}(\mathbf{x})$ is a diagonal matrix containing the variances of each global process variation parameter. Subsequently, conditional on $\mathbf{h}^{(1)}$, local mismatch parameters $\mathbf{h}^{(2)}$ are sampled from $\mathcal{N}\big(\mathbf{h}^{(1)}, \Sigma^{\text{Local}}(\mathbf{x})\big)$. Here, $\Sigma^{\text{Local}}(\mathbf{x})$ is a diagonal matrix containing the variances of the device-specific variations, which depend on the design solution $\mathbf{x}$ \cite{device_mismatch}. Finally, the mismatch condition $\mathbf{h}$ is sampled by combining global and local variations for given design configuration \cite{process_variation2, hspice_manual}.

\begin{table}[t]
\caption{Operational Configuration of the Framework}
\centering
\setlength{\extrarowheight}{3pt}
\resizebox{\linewidth}{!}{
\begin{tabular}{C{1.2cm}|C{0.5cm}|C{0.5cm}|C{0.5cm}|c|c|c|c}  % Set width to columnwidth
\Xhline{2.3\arrayrulewidth} % 위쪽 두꺼운 선
Verif. & \multicolumn{3}{c|}{Predefined Corner $\mathbf{t}$} & \multicolumn{2}{c|}{Var. of Mismatch $\mathbf{h}$} & \multicolumn{2}{c}{\# of Samples} \\ \cline{2-8} 
Method            & P & V & T & Global & Local & Optim. & Verif. \\ \hline
C                 & Y & Y & Y & $0$ & $0$ & $1$ & $k$ \\ \hline
C-MC\textsubscript{L} & Y & Y & Y & $0$ & $\Sigma^{Local}$ & $N'$ & $k\times N$  \\ \hline
C-MC\textsubscript{G-L} & N & Y & Y & $\Sigma^{Global}$ & $\Sigma^{Local}$ & $N'$   & $k\times N$  \\ \Xhline{2.3\arrayrulewidth}
\end{tabular}
}
\vspace{-0.4cm}
\end{table}

\subsection{Operational Configuration}
The GLOVA framework provides flexibility in selecting the target verification method (e.g., corner simulation, MC simulation, or a combination of both). With this adaptability, GLOVA efficiently adjusts the sampling methods and quantity during the optimization process. The configuration of the operational variables based on the chosen verification method is summarized in Table I.

\textbf{C: Corner simulation.} Predefined corners are considered without including mismatch conditions.

\textbf{C-MC\textsubscript{L}: Corner and local MC simulation.} Mismatch conditions are sampled from the local variation distribution for predefined corners.

\textbf{C-MC\textsubscript{G-L}: Corner and global-local MC simulation.} Mismatch conditions are sampled from the combined distribution of global and local variations for predefined corners.

% For both C-M\textsubscript{L} and C-M\textsubscript{G-L}, a small number of $N’$ mismatch conditions are sampled during the optimization phase, while a large number $N$ are sampled across all given corners during the verification phase.

\subsection{Overall Workflow}
An overview of the GLOVA framework is provided in Fig. 2. Initially, it utilizes TuRBO \cite{turbo} to generate design solutions that meet constraints under the typical condition. This initial sampling is adopted from \cite{pvt_sizing}. Each design solution is then simulated across sampled mismatch conditions under all PVT corners. The worst-case reward from these simulations is stored in the worst-case replay buffer, forming the initial dataset. Additionally, the last worst-case buffer records the last worst reward of each PVT corner. In each iteration, the following steps are executed.
\blackcircle{1} \textbf{Generate a design solution.} Put the last design solution into an actor to get a new design solution.
\blackcircle{2} \textbf{Sample PVT conditions.} Select the worst PVT corner by comparing the last worst-cases of corners, and sample $N'$ mismatch conditions from the distribution via Eq. (3).
\blackcircle{3} \textbf{Simulate the design solution under the sampled conditions.} Get performance metrics under the sampled $N'$ mismatch conditions given the worst PVT corner.
\blackcircle{4} \textbf{Evaluate the design solution via $\mu$-$\sigma$ metric. If it is decided not to verify further, go to Step 6.}
\blackcircle{5} \textbf{Fully verify with reordered PVT conditions.} Simulate the design solution under the targeted verification method. If the design meets all constraints in all cases, the framework terminates. Otherwise, continue the optimization process.
\blackcircle{6} \textbf{Store the worst reward in the replay buffer and update RL agent with data stored in the buffer.} Only the worst reward is stored across PVT conditions.

\section{Optimization Phase of Proposed GLOVA}

\subsection{Risk-Sensitive Reinforcement Learning Agent}

Risk-sensitive RL \cite{risk_survey} is an approach that accounts for stochastic uncertainty within a system. To avoid uncertain and costly failures, agents are trained to maximize the worst-case robustness of the reward. This risk-sensitivity potentially leads to faster learning for unpredictable systems. In GLOVA, a risk-sensitive RL agent is employed to find the desired design solution in a situation where performance metrics fluctuate by various PVT conditions. By treating each variation as a potential risk and adopting a risk-avoidance policy, this approach reduces the cost of numerous simulations required by industrial-level verification methods, such as corner and MC simulations.

\textbf{Reward.}
The actor-critic agent, which is a widely used method in RL \cite{actor_critic}, is trained to optimize the multiple performance metrics to meet constraints. To simplify this process, we use a reward function that consolidates multiple objectives into a single target. The reward is defined as follows:

\vspace{-0.3cm}
\begin{equation}
r =
\begin{cases}
r', & r' < 0 \\
0.2, & r' \geq 0
\end{cases}
\end{equation}

\noindent with \(r'\) calculated as:

\vspace{-0.3cm}
\begin{equation}
r' = \sum_{i=1}^{m} \min \left( f_i, 0 \right)
\end{equation}

\noindent where $f_i$ is the normalized current simulated $i^{th}$ performance metric, defined as $f_i = (c_i - \mathcal{F}_i) / (c_i + \mathcal{F}_i)$, and $c_i$ is the corresponding constraint. As we want $f_i \leq c_i$, a smaller reward indicates a worse design. If all constraints are satisfied, the reward is set to $0.2$. This reward formulation is modified from \cite{pvt_sizing, robust_analog}.

\textbf{Actor and critic.}
The actor is a 4-layer neural network. The actor’s input is the previous design $\mathbf{x}_{last}$, a $p$-dimensional normalized design solution, where each dimension represents a design parameter (e.g. width or length in a transistor). The output of the actor is the next design $\mathbf{x}_{new}$, a $p$-dimensional normalized design solution. On the other hand, the critic includes a set of 4-layer neural networks, which are called base models. The input of the critic is also a $p$-dimensional normalized design solution. The critic outputs a scalar value representing the input's predicted reliability bound under various PVT conditions. The process of predicting uncertain design reliability bounds using the base models is detailed in Section IV. B.

\begin{algorithm}[t]
\caption{Risk-sensitive RL in GLOVA}
    Given replay buffer $B^{worst}$ and the last design $\mathbf{x}_{last}$\; 
    Given actor network \(A(\mathbf{x}|\theta^A)\) and critic network \(Q(\mathbf{x}|\theta^Q)\) with a set of base models \(\{Q_i(\mathbf{x}|\theta^Q_i)\}\)\;
    Given the number of samples $N'$ and the worst corner $\mathbf{t}^{worst}$\;
    
    \For{iteration = 1, $M$}{
        Sample a batch of $(\hat{\mathbf{x}}, \hat{r})$ from \(B^{worst}\)\;
        \For{i = 1, ensemble size}{
            Update the base model by minimizing the loss:
            \\$\mathcal{L}_{Q_i} = MSELoss(\hat{r}, Q_i(\hat{\mathbf{x}}|\theta^Q_i) + bias)$\;}
        Update the actor by minimizing the loss:
        $\mathcal{L}_{A} = MSELoss(0.2, Q(A(\hat{\mathbf{x}}|\theta^A)|\theta^Q) + bias)$\;
        Select a new design according to the current policy and exploration noise: \(\mathbf{x}_{new}=A(\mathbf{x}_{last}|\theta^A)+noise\)\;
        Sample mismatch conditions \(\widetilde{H}^{N'}\) via Eq. (3)\;
        Simulate the \(\{\mathbf{x}_{new}|\mathbf{t}^{worst}, \widetilde{H}^{N'}\}\) to get rewards $\{r\}$\;
        Select the worst-case reward: $r^{worst}=\min \{r\}$\;
        Store the data (\(\mathbf{x}_{new}\),\(r^{worst}\)) in $B^{worst}$\;
        }
\end{algorithm}

\textbf{Training.}
Algorithm 1 outlines agent’s training procedure. Here, $M$ is the maximum number of optimization iterations. The training process is modified from DDPG \cite{DDPG}. Unlike risk-neutral training, the risk-avoidance process evaluates each design iteration based on worst-case scenarios under sampled PVT conditions. In other words, while multiple rewards are obtained from simulations, only the worst reward is stored in the replay buffer and used for training. Each of the critic’s base models is independently trained with a distinct batch sampled from the replay buffer. The actor is trained to find the desired design solution using the predicted reliability bound, which is derived by aggregating the outputs of the critic’s base models.

\subsection{Ensemble-based Critic}

\begin{figure}[t]
  \centering
  \includegraphics[width=\linewidth]{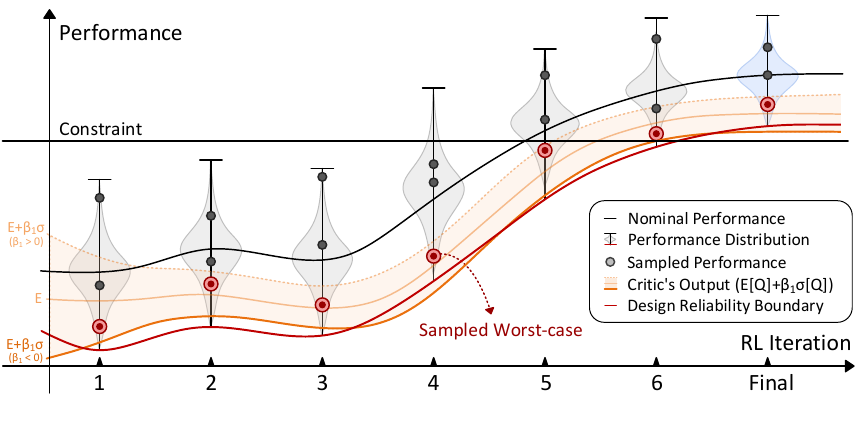}
  \vspace{-0.8cm}
  \caption{Design reliability bound estimation under PVT variations using an ensemble-based critic trained on sampled worst-case scenarios.}
  \label{Fig:3}
  \vspace{-0.5cm}
\end{figure}

In risk-sensitive RL, modeling worst-case scenarios is crucial, yet impractical due to the need for over 1,000 samples per iteration to determine reliability bounds under PVT variations \cite{mc1, mc2}. Therefore, to achieve sample-efficient risk management, we propose an ensemble-based critic that approximates design reliability bounds for variation by sampling worst cases. This approach allows the actor to explore a wide range of variations without simulating each scenario individually, improving sample-efficiency.

The proposed critic extends the risk-sensitive criterion \cite{risk_survey} by employing an ensemble of base models. Fig. 3 qualitatively illustrates the modeling of the design reliability boundary with the ensemble-based critic during the optimization process. First, in each iteration, a small number $N'$ (typically 2 to 5) of mismatch conditions $\widetilde{H}^{N'}$ are sampled and simulated under the worst-case corner to capture performance variability. The critic then learns from the worst-case among $N'$ sampled variations, using an ensemble to estimate design reliability bound with awareness of uncertainty. Each base model is trained with different data batches, benefiting from randomness and varying initialization. This process allows the critic to compensate for the uncertainty in the design reliability bound caused by limited statistical samples. The output of critic is defined as follows:

\vspace{-0.2cm}
\begin{equation}
    Q(\mathbf{x}|\theta^Q) = \mathbb{E}[Q_i(\mathbf{x}|\theta^Q_i)]+\beta_1\sigma[Q_i(\mathbf{x}|\theta^Q_i)]
\end{equation}

\noindent where $Q_i(\mathbf{x}|\theta^Q_i)$ represents the output of the $i^{th}$ base model with weights $\theta^Q_i$, $\mathbb{E}[\cdot]$ denotes the average of the base model outputs, $\sigma[\cdot]$ refers to the standard deviation of the base model outputs, and $\beta_1 < 0$ is the risk-avoidance parameter. The critic manages risk and provides reliability bounds to guide the actor in finding a feasible design solution.

\section{Verification Phase of Proposed GLOVA}
Extensive simulations, which are time-intensive, are inevitable to ensure circuit reliability under PVT variations. If a design fails to meet its target during the verification process, a large number of simulations can become an inefficient use of resources. Therefore, to achieve sample-efficient verification, it is crucial to detect failures early and halt the verification process, returning to the optimization phase. To address this, we propose a hierarchical verification algorithm comprising the $\mu$-$\sigma$ evaluation and the simulation reordering method. The verification workflow of GLOVA is presented in Algorithm 2.

\subsection{$\mu$-$\sigma$ Evaluation Method}
The $\mu$-$\sigma$ evaluation first analyzes a subset $N’$ of the total $N$ MC simulations under the given corner conditions. Based on this analysis, it determines whether it is worthwhile to proceed with the remaining $N - N’$ simulations for full verification. Specifically, it statistically estimates the performance distribution of the full mismatch condition set $\widetilde{H}^{N}$ using the pre-sampled subset $\widetilde{H}^{N'}$. The $\mu$-$\sigma$ evaluation is conducted sequentially across all given corners, starting with the worst corner from the last worst-case buffer. Note that, during the optimization phase, the $\widetilde{H}^{N’}$ for the worst corner has already been simulated and can be reused. If the evaluation fails to pass the criteria set in Eq. (7), the design is deemed to have failed verification. The evaluation criterion is provided by the following equation:

\vspace{-0.2cm}
\begin{equation}
e_i = \mathbb{E}[f_i] + \beta_2\sigma[f_i] \leq c_i
\end{equation}

\noindent where $f_i$ represents the $i^{th}$ normalized performance metric and $\beta_2$ is a reliability factor. The reliability factor $\beta_2$ is set to a positive value of $4$ or higher, as higher values in the performance metric, unlike rewards, indicate worse performance. This reliability factor compensates for the incomplete nature of the distribution caused by a lack of samples. The $\mu$-$\sigma$ evaluation method conservatively assesses whether a given design is feasible for full verification, reducing the likelihood of verification attempts that would ultimately result in failure. Consequently, this approach saves valuable resources and reduces overall runtime.

\begin{algorithm}[t]
    \caption{Verification Algorithm of GLOVA}
    Given the number of samples $N$ for full verification and a subset of samples $N’$\;
    Sort $T$ based on the last worst-case buffer\;
    \For{each $\mathbf{t}_j$ in sorted $T$}{
        Sample \(\widetilde{H}^{N'}_{j}\) from the distribution via Eq. (3)\;
        Simulate \(\{\mathbf{x}|\mathbf{t}_j, \widetilde{H}^{N'}_j\}\) to obtain \(\{r,f_i(\mathbf{x}|\mathbf{t}_j,\widetilde{H}^{N'}_j)\}\)\;
        \If{$\mu$-$\sigma$ evaluation passes}{
            Calculate \(\texttt{t-SCORE}_{j}\)\;
            Calculate Pearson correlation coefficient vector \(\rho_j\)\;
        }
        \Else{
            Verification failed\;
        }
    }

    Sort $T$ by \(\{\texttt{t-SCORE}_{j}\}\)\;
    
    \For{each $\mathbf{t}_j$ in sorted $T$}{
        Sample \(\widetilde{H}^{N-N'}_j\) from the distribution via Eq. (3)\;
        Calculate \(\texttt{h-SCORE}_{j,n}\) for each \(\mathbf{h}_{j,n} \in \widetilde{H}^{N-N'}_j\) and \(\rho_j\)\;
        Sort \(\widetilde{H}^{N-N'}_j\) by \(\{\texttt{h-SCORE}_{n}\}_j\)\;

        \For{each $\mathbf{h}_{j,n}$ in sorted \(\widetilde{H}^{N-N'}_j\)}{
            Simulate \(\{\mathbf{x}|\mathbf{t}_j, \mathbf{h}_{j,n}\}\) to obtain \(\{r,f_i(\mathbf{x}|\mathbf{t}_j,\mathbf{h}_{j,n})\}\)\;
            \If{$r \neq 0.2$}{
                Verification failed\;
            }
        }
    }
\end{algorithm}

\begin{figure*}[t]
  \centering
  \includegraphics[width=\textwidth]{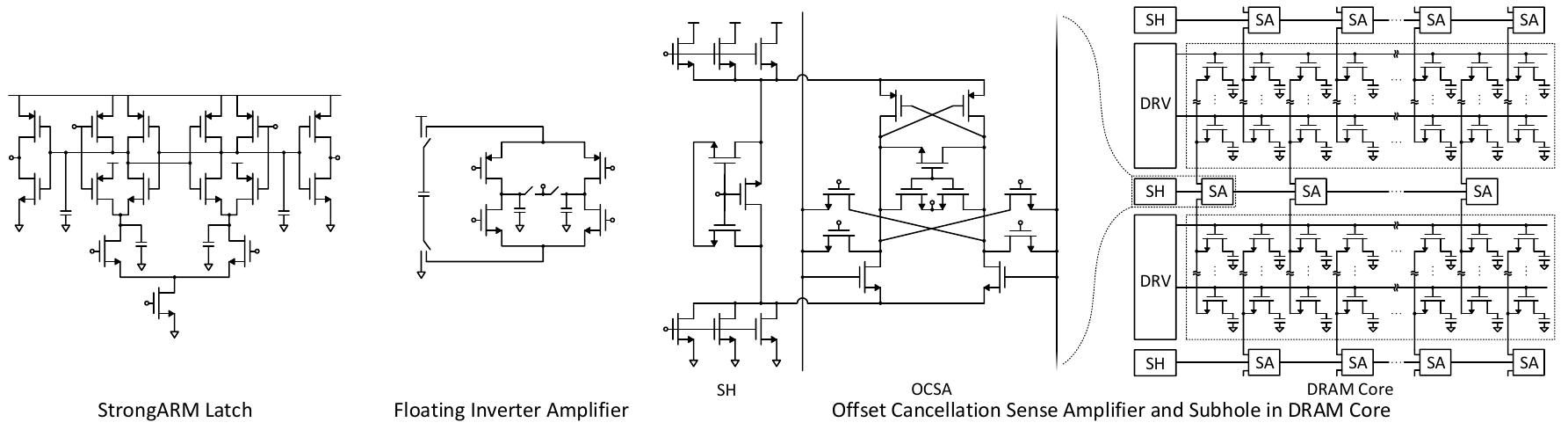}
  \vspace{-0.7cm}
  \caption{Schematics of three analog/mixed-signal testcase circuits.}
  \label{Fig:4}
  \vspace{-0.3cm}
\end{figure*}

\subsection{Simulation Reordering Method}

If the design passes the $\mu$-$\sigma$ evaluation for the $N’$ samples across all given corners, full verification is performed for the remaining $N-N’$ samples for each corner. To detect failure early and halt the verification process, the simulation sequence is determined by the proposed corner and MC reordering methods, prioritizing simulations with a higher likelihood of failure.

\textbf{Corner reordering.} For the pre-sampled mismatch condition set $\widetilde{H}^{N’}_j$ of each corner $\mathbf{t}_j$, the $\texttt{t-SCORE}_j$ is calculated to rank the severity of degradation caused by mismatch conditions as follows:
\vspace{-0.1cm}
\begin{equation}
\texttt{t-SCORE}_{j} = \sum^{m}_{i=1} e_{j,i}
\end{equation}

\noindent where $e_{i}$ is computed by Eq. (7). Corners with a higher $\texttt{t-SCORE}$, indicating a greater potential for failure, are selected for simulation first.

\textbf{MC reordering.} For the given corner, the simulation list for the remaining mismatch condition set $\widetilde{H}^{N-N’}_j$ is determined. To prioritize the mismatch conditions most likely to result in failure, the Pearson correlation coefficient is introduced to assess the relationship between the mismatch parameters and performance. Specifically, for each \( \mathbf{h}_{j,n} \) vector in the pre-sampled set \( \widetilde{H}^{N'}_j \), the Pearson correlation coefficient vector \( \rho_j \) is calculated between each element of \( \mathbf{h}_{j,n} \) and the corresponding performance metric \( g_{j,n} \), where \( g = \sum_i f_i \), as defined by the following equation:

\begin{equation}
\rho_j = \frac{\sum_{n=1}^{N'} (\mathbf{h}_{j,n} - \bar{\mathbf{h}_j})(g_{j,n} - \bar{g_j})}{\sqrt{\sum_{n=1}^{N'} (\mathbf{h}_{j,n} - \bar{\mathbf{h}_j})^2} \sqrt{\sum_{i=1}^{N'} (g_{j,n} - \bar{g_j})^2}}.
\end{equation}

\noindent Then, the $\texttt{h-SCORE}_{j,n}$ is defined as the weighted sum of the mismatch condition vector $\mathbf{h}_{j,n}$ from the mismatch condition set $\widetilde{H}^{N-N'}_j$ and correlation vector $\rho_j$, as follows:

\begin{equation}
\texttt{h-SCORE}_{j,n} = \sum_{i=1}^{r} \left[(\mathbf{h}_{j,n})_i \circ (\rho_j)_i\right].
\end{equation}

\noindent Mismatch conditions with a higher $\texttt{h-SCORE}$, indicating a greater potential for failure, are prioritized for simulation for the given corner.

By reordering both the corners and mismatch conditions, failures in the verification process can be detected early, effectively reducing the number of unnecessary simulations. This simulation reordering method reduces the verification cost and improves the overall efficiency of the framework.

\section{Results}

\subsection{Analog/mixed-signal Circuits}
Real-world designs from previous work \cite{robust_analog}, \cite{pvt_sizing} are adopted as testcases to compare the design efficiency of GLOVA. The testcases include the strongARM latch (SAL) \cite{SAL} and the floating inverter amplifier (FIA) \cite{FIA}.  These circuits are selected due to their fully dynamic operation, which makes them highly sensitive to PVT variations. Additionally, we include the offset cancellation sense amplifier (OCSA) \cite{OCSA_original} and subhole (SH) in DRAM core with an $6\text{F}^2$ open bitline architecture, which consists of 2K wordlines and peripherals for memory operation \cite{OCSA}. This testcase is particularly challenging due to large parasitic array capacitance and extensive mismatches, which necessitate a significantly higher number of statistical simulations to achieve a high yield. The topologies are presented in Fig. 4.
% Note that these circuits are selected due to their fully dynamic operation, which makes them highly sensitive to PVT variations and suffer from extensive mismatches. The topologies are presented in Fig. 4.

All testcases are designed using advanced 28$\mathrm{n\/m}$ CMOS technology and simulated with a SPICE-based simulator \cite{hspice_manual} under 30 PVT conditions, given by $\{\mathrm{TT}, \mathrm{SS}, \mathrm{FF}, \mathrm{SF}, \mathrm{FS}\} \times \{0.8\mathrm{V}, 0.9\mathrm{V}\} \times \{-40\thinspace\mathrm{^\circ \/C}, 27\thinspace\mathrm{^\circ \/C}, 80\thinspace\mathrm{^\circ \/C}\}$. Each circuit comprises multiple transistors and capacitors, where the transistors are defined by parameters such as gate width and length, and the capacitors by their capacitance values. Mismatch parameters are considered for each device, with variances following the PDK rules for the same technology.

\textbf{StrongARM latch.}
The sizing vector of this circuit consists of 14 parameters: 6 transistor widths, 6 transistor lengths, and 2 capacitances. The range for each parameter is \([0.28, 32.8] \, \mathrm{\mu \/m}\) for width, \([0.03, 0.33] \, \mathrm{\mu \/m}\) for length, and \([0.005, 5.5] \, \mathrm{p\/F}\) for capacitance with a total design space of \(10^{28}\). The mismatch parameter includes all devices in the circuit. The performance metrics are power, set delay, reset delay, and noise. The design targets are as follows, which are the same as \cite{pvt_sizing}:

\vspace{-0.4cm}
% \[
% c = 
% \left\{
% \begin{align*}
% \begin{array}{r@{\hskip}l}
%     \text{Power } &\leq 40 \thinspace\mathrm{\mu\/W} \\[1mm]
%     \text{Set delay } &\leq 4 \thinspace\mathrm{n\/s} \\[1mm]
%     \text{Reset delay } &\leq 4 \thinspace\mathrm{n\/s} \\[1mm]
%     \text{Noise } &\leq 120 \thinspace\mathrm{\mu\/V}.
% \end{array}
% \right.
% \end{align*}
% \]

\[
c = 
\left\{
\begin{array}{r@{\hskip 0.5em}l}
    \text{Power}        &\leq 40\,\mathrm{\mu W} \\[1mm]
    \text{Set delay}    &\leq 4\,\mathrm{ns} \\[1mm]
    \text{Reset delay}  &\leq 4\,\mathrm{ns} \\[1mm]
    \text{Noise}        &\leq 120\,\mathrm{\mu V}
\end{array}
\right.
\]

\textbf{Floating inverter amplifier.}
The sizing vector of this circuit consists of 6 parameters: 2 transistor widths, 2 transistor lengths, and 2 capacitances. The range for each parameter is the same as those of the strongARM latch with a total design space of \(10^{12}\). The mismatch parameter includes all devices in the circuit. The performance metrics are energy consumption per conversion and noise. The design targets are as follows, taking technology scaling into account in \cite{pvt_sizing}:
% \[
% c = 
% \left\{
% \begin{align*}
% \begin{array}{r@{\hskip}l}
%     \text{Energy/conv. } &\leq 0.1 \thinspace\mathrm{p\/J} \\[1mm]
%     \text{Noise } &\leq 130 \thinspace\mathrm{m\/V}. \\
% \end{array}
% \right.
% \end{align*}
% \]
\[
c = 
\left\{
\begin{array}{r@{\hskip 0.5em}l}
    \text{Energy/conv.} &\leq 0.1\,\mathrm{pJ} \\[1mm]
    \text{Noise} &\leq 130\,\mathrm{mV}
\end{array}
\right.
\]

\begin{table*}[ht]
\centering
\normalsize
\caption{Optimization Results on Real-World Circuits}
\label{tab:2}
\setlength{\extrarowheight}{1pt}
\setlength{\tabcolsep}{8pt}
\resizebox{\textwidth}{!}{%
\begin{tabular}{cc|C{1.2cm}C{1.2cm}C{1.2cm}|C{1.2cm}C{1.2cm}C{1.2cm}|C{1.2cm}C{1.2cm}C{1.2cm}} \Xhline{2.3\arrayrulewidth}
\multicolumn{2}{c|}{Testcases} & \multicolumn{3}{c|}{SAL} & \multicolumn{3}{c|}{FIA} & \multicolumn{3}{c}{OCSA and SH in DRAM Core} \\ \hline
\multicolumn{2}{c|}{Verification} & C & C-MC\textsubscript{L} & C-MC\textsubscript{G-L} & C & C-MC\textsubscript{L} & C-MC\textsubscript{G-L} & C & C-MC\textsubscript{L} & C-MC\textsubscript{G-L}  \\ \hline
\multirow{3}{*}{RL Iteration} & Ours & 6 & 8 & 12 & 18 & 26 & 48 & 21 & 84 & 129 \\
 & PVTSizing & 19 & 24 & 27 & 48 & 71 & 138 & 72 & 138 & 238$^*$ \\
 & RobustAnalog & 104 & 124 & 297 & 533 & 840$^*$ & 1,733$^*$ & 760 & 1,166$^*$ & 2,064$^*$ \\ \hline
\multirow{3}{*}{\# Simulation} & Ours & 83 & 3,103 & 8,809 & 248 & 3,203 & 6,461 & 390 & 6,916 & 72,853 \\
 & PVTSizing & 186 & 10,748 & 31,221 & 322 & 87,773 & 293,076 & 2,066 & 300,332 & 224,768$^*$ \\
 & RobustAnalog & 442 & 12,683 & 75,920 & 2,151 & 146,889$^*$ & 361,066$^*$ & 6,406 & 557,050$^*$ & 753,048$^*$ \\ \hline
\multirow{3}{*}{Norm. Runtime} & Ours & 1.00 & 1.00 & 1.00 & 1.00 & 1.00 & 1.00 & 1.00 & 1.00 & 1.00 \\
 & PVTSizing & 2.77 & 3.45 & 3.81 & 1.71 & 26.28 & 43.53 & 3.85 & 40.59 & 3.07$^*$ \\
 & RobustAnalog & 11.17 & 4.43 & 9.63 & 14.94 & 45.26$^*$ & 55.02$^*$ & 21.24 & 76.03$^*$ & 10.40$^*$ \\ \hline
\multirow{3}{*}{Success Rate} & Ours & 100\% & 100\% & 100\% & 100\% & 100\% & 100\% & 100\% & 100\% & 100\% \\
 & PVTSizing & 100\% & 100\% & 100\% & 100\% & 100\% & 100\% & 100\% & 100\% & 87\% \\
 & RobustAnalog & 100\% & 100\% & 100\% & 100\% & 95\% & 90\% & 100\% & 83\% & 53\% \\ \Xhline{2.3\arrayrulewidth}
\end{tabular}%
}
\begin{flushleft}
\footnotesize
$^*$In tests where the success rate is below 100\%, only data from successful optimizations are included.
\end{flushleft}
\vspace{-0.4cm}
\end{table*}

\begin{table}[ht]
\centering
\caption{Ablation Study}
\label{tab:3}
\setlength{\extrarowheight}{1pt}
\setlength{\tabcolsep}{8pt}
\resizebox{\linewidth}{!}{%
\begin{tabular}{cc|C{1.1cm}C{1.1cm}C{1.1cm}} \Xhline{2.1\arrayrulewidth}
\multicolumn{2}{c|}{Verification} & C & C-MC\textsubscript{L} & C-MC\textsubscript{G-L} \\ \hline
\multirow{4}{*}{RL Iteration} & Proposed & 21 & 84 & 129 \\
 & w/o EC$^1$ & 26 & 92 & 199$^*$ \\
 & w/o $\mu$-$\sigma$$^2$ & - & 101 & 239$^*$ \\
 & w/o SR$^3$ & - & - & - \\
 \hline
\multirow{4}{*}{\# Simulation} & Proposed & 390 & 6,916 & 72,853 \\
 & w/o EC & 1,218 & 18,232 & 212,153$^*$ \\
 & w/o $\mu$-$\sigma$ & - & 136,217 & 476,721$^*$ \\
 & w/o SR & 2,448 & 253,738 & 765,375$^*$ \\ \hline
 \multirow{4}{*}{Norm. Runtime} & Proposed & 1.00 & 1.00 & 1.00 \\
 & w/o EC & 3.75 & 3.02 & 3.32$^*$\\
 & w/o $\mu$-$\sigma$ & - & 21.97 & 7.45$^*$ \\
 & w/o SR & 7.05 & 40.80 & 11.93$^*$ \\ \hline
\multirow{4}{*}{Success Rate} & Proposed & 100\% & 100\% & 100\% \\
 & w/o EC & 100\% & 100\% & 90\% \\
 & w/o $\mu$-$\sigma$ & 100\% & 100\% & 95\% \\
 & w/o SR & 100\% & 100\% & 95\% \\ \Xhline{2.1\arrayrulewidth}
\end{tabular}%
}
\begin{flushleft}
$^*$In tests where the success rate is below 100\%, only data from successful optimizations are included.\\
$^1$Ensemble-based critic, $^2$$\mu$-$\sigma$ evaluation, $^3$Simulation reordering.
\end{flushleft}
\vspace{-0.4cm}
\end{table}

\textbf{Offset cancellation sense amplifier and subhole in DRAM core.} The sizing vector of this circuit consists of 12 parameters: 6 transistor widths and 6 transistor lengths. The range for each parameter is \([0.28, 1.028] \, \mu \mathrm{m}\) for the transistor width in the OCSA, \([5, 15] \, \mu \mathrm{m}\) for the transistor width in the SH, and \([0.03, 0.06] \, \mu \mathrm{m}\) for all transistor length, resulting in a total design space of \(10^{24}\). These parameter ranges are determined with consideration for cell pitch, as the transistors need to be integrated near the cell array. The mismatch parameter includes all devices in the DRAM core. The simulation methods and metrics are based on \cite{OCSA}. The performance metrics are low data sensing voltage \(\Delta V_{D0}\), high data sensing voltage \(\Delta V_{D1}\), and energy consumption per 1-bit sensing. Note that since both the low and high data sensing voltages are metrics that need to be maximized, their signs are inverted. The design targets are as follows:

\vspace{-0.3cm}
% \[
% c = 
% \left\{
% \begin{align*}
% \begin{array}{r@{\hskip}l}
% \Delta V_{D0}\text{ } &\geq 85 \thinspace\mathrm{m\/V}\text{ } \rightarrow -\Delta V_{D0} \leq -85 \thinspace\mathrm{m\/V} \\[1mm]
% \Delta V_{D1}\text{ }  &\geq 85 \thinspace\mathrm{m\/V}\text{ } \rightarrow -\Delta V_{D1} \leq -85 \thinspace\mathrm{m\/V}\\[1mm]
% \text{Energy/bit }  &\leq 30 \thinspace\mathrm{f\/J}. \\
% \end{array}
% \right.
% \end{align*}
% \]

\[
c = 
\left\{
\begin{array}{r@{\hskip 0.5em}l}
\Delta V_{D0} &\geq 85\,\mathrm{mV} \rightarrow -\Delta V_{D0} \leq -85\,\mathrm{mV} \\[1mm]
\Delta V_{D1} &\geq 85\,\mathrm{mV} \rightarrow -\Delta V_{D1} \leq -85\,\mathrm{mV} \\[1mm]
\text{Energy/bit} &\leq 30\,\mathrm{fJ}
\end{array}
\right.
\]

\subsection{Evaluation and Analysis}
To validate GLOVA’s design solution across various PVT corner conditions and extensive mismatches, we incorporate three verification scenarios: 30 PVT corner simulations (C), 0.1K local Monte Carlo simulations on 30 PVT corners (C-MC\textsubscript{L}), and 1K global-local Monte Carlo simulations on 6 VT corners (C-MC\textsubscript{G-L}). Each method requires 30, 3,000, and 6,000 simulations, respectively, to complete full verification. In our setting, simulations are conducted in parallel with a sample size of 3 during the optimization phase, while the verification phase utilizes the maximum available resources. For training, the batch size is set to 10, and the risk-avoidance parameter $\beta_1$ and reliability factor $\beta_2$ are set to -3 and 4, respectively.

The optimization results across three real-world circuits are shown in Table II. GLOVA achieves the highest success rate, with up to 80.5$\times$ greater sample efficiency and 76.0$\times$ lower time consumption compared to PVTSizing and RobustAnalog. Notably, for the OCSA and SH in the DRAM core case, which is verified using the C-MC\textsubscript{G-L} method, GLOVA demonstrates 1.9$\times$ improvement in success rate and 10.3$\times$ improvement in sample efficiency over PVTSizing and RobustAnalog. However, this case appears to consume more RL iterations and simulations than other cases. This is due to the challenge of designing a solution that simultaneously satisfies low data sensing voltage and high data sensing voltage—two conflicting metrics—in scenarios where the DRAM core circuit is highly sensitive to mismatch cases across numerous devices within the cell array.

Table III presents the results of an ablation study conducted on a DRAM core to assess the contributions of the proposed methods. The ensemble-based critic facilitates the derivation of feasible design solutions during the optimization process, improving in both success rate and sample efficiency. Additionally, the $\mu$-$\sigma$ evaluation and simulation reordering methods effectively reduce the number of simulations required in the verification process, thereby lowering verification costs and enhancing the overall efficiency of the framework.

\section{Conclusion}
We present GLOVA, an efficient optimization and verification framework that effectively manages diverse PVT conditions and extensive mismatches. GLOVA employs risk-sensitive reinforcement learning and introduces an ensemble-based critic for variation-aware optimization. Incorporating $\mu$-$\sigma$ evaluation and simulation reordering methods significantly alleviates the simulation burden during verification, improving the overall efficiency of the framework. The proposed GLOVA supports industrial-level corner and Monte Carlo simulations for design verification. Experimental results on real-world circuits, including challenging DRAM core, demonstrate that GLOVA substantially reduces simulation and time costs compared to previous state-of-the-art frameworks.

\section*{Acknowledgment}

\end{document}